# Generic Tracking and Probabilistic Prediction Framework and Its Application in Autonomous Driving

Jiachen Li, *Student Member, IEEE*, Wei Zhan, *Student Member, IEEE*, Yeping Hu, *Student Member, IEEE*, and Masayoshi Tomizuka, *Life Fellow, IEEE*

*Abstract*—Accurately tracking and predicting behaviors of surrounding objects are key prerequisites for intelligent systems such as autonomous vehicles to achieve safe and high-quality decision making and motion planning. However, there still remain challenges for multi-target tracking due to object number fluctuation and occlusion. To overcome these challenges, we propose a constrained mixture sequential Monte Carlo (CMSMC) method in which a mixture representation is incorporated in the estimated posterior distribution to maintain multi-modality. Multiple targets can be tracked simultaneously within a unified framework without explicit data association between observations and tracking targets. The framework can incorporate an arbitrary prediction model as the implicit proposal distribution of the CMSMC method. An example in this paper is a learning-based model for hierarchical time-series prediction, which consists of a behavior recognition module and a state evolution module. Both modules in the proposed model are generic and flexible so as to be applied to a class of time-series prediction problems where behaviors can be separated into different levels. Finally, the proposed framework is applied to a numerical case study as well as a task of on-road vehicle tracking, behavior recognition, and prediction in highway scenarios. Instead of only focusing on forecasting trajectory of a single entity, we jointly predict continuous motions for interactive entities simultaneously. The proposed approaches are evaluated from multiple aspects, which demonstrate great potential for intelligent vehicular systems and traffic surveillance systems.

## I. INTRODUCTION

EFFECTIVE tracking of surrounding objects and inference of their future motions are critical for intelligent systems (e.g. autonomous vehicles and industrial robotics) to achieve safe and high-quality decision making, motion planning and control. Although single-target tracking problems have been well studied in literature, it still remains a challenge for multi-target tracking due to multi-modality and selection of data association methods. The tracking performance is also dependent on the prediction quality of state evolution models. In simple scenarios where each entity behaves independently or with few interactions, state transition models based on pure kinematics or dynamics can be accurate enough to make a short-term forecasting. In many real-world applications where highly-interactive agents exist, however, these models are not sufficient due to the inherent uncertainty of future behaviors and interdependency among multiple entities. Also these models may be even not available due to the essential complexity such as pedestrian behaviors. Above all, it is desired to have a unified framework for tracking multiple agents and predicting their future motions simultaneously, which takes the uncertainties and interactions into account.

Many studies on multi-target tracking have been conducted in recent decades, which can be classified into two main categories. One category employs deep learning and computer vision techniques to track objects by real-time detection on images and videos [1], [2], where the bounding boxes of tracking targets can be obtained. The other category estimates the state distribution of tracking targets through Bayesian inference methods. Kalman filter (KF) [3] is a widely used estimator for linear systems while Extended Kalman filter (EKF) [4], [5] and Unscented Kalman filter (UKF) [6] are utilized in nonlinear systems. However, in practice the state distribution cannot be well approximated by simple multivariate Gaussian distribution. The sequential Monte Carlo (SMC) method (a.k.a. particle filter) [7]–[10] thus has superiority over KF variants since no assumption on system model and state distribution is made. There are also research lying in the intersection of both categories [9], in which Bayesian filters are utilized to provide a heuristic for forecasting the positions of bounding boxes in future frames. In this paper, we only focus on tracking methods based on recursive Bayesian state estimation where observation sequences can be obtained by sensor fusion.

There are two popular solutions for multi-target tracking with SMC method. One is using multiple instances of single object tracking where each entity is assumed to be independent. However, independence is not an appropriate assumption for interactive targets. The other is using dynamic state space extension to model the joint distribution of all the objects' states. However, the dimension of state space will blow up as target number increases. To overcome the challenges and deficiencies of existing methods, we propose a uniform framework which makes a bridge for multi-target tracking and multi-agent prediction. Instead of assuming each tracking target behaves independently, we take the interactions into consideration to enhance tracking.

There are also many research efforts devoted to improve the performance of behavior and motion prediction which can be sorted as three main classes in terms of prediction result representation. The first class [11], [12] provides one or a group of possible future trajectories to represent motion patterns which are defined as "prototype trajectory" in [13]. The second class gives continuous actions such as velocities, accelerations and yaw angles of vehicles at each time step [14]–[16]. The third class disregards the identity of each entity and employ an occupancy grid map to provide a uniform representation [10], [17]. The prediction approaches can also be classified into deterministic and probabilistic prediction. Although many studies have demonstrated satisfactory performance of deterministic methods in less interactive scenarios, considering uncertainties is significant implying that probabilistic methods are more superior in scenarios full of uncertainties. More related research are provided in Section II.

The main contributions of this paper are summarized as follows: (a) We propose a constrained mixture sequential Monte Carlo (CMSMC) method in which a mixture representation is incorporated in the estimated posterior distribution to maintain multi-modality. Multiple targets can be tracked simultaneously within a unified framework without explicit data association between observations and tracking targets. Any form of prediction models can be employed as the implicit proposal distribution of CMSMC method. (b) We also put forward a learning-based model as an instance for hierarchical time-series prediction, which consists of a behavior recognition module and a state evolution module. Both modules in the proposed model are generic and flexible so as to be applied to a class of time-series prediction problems where behaviors can be separated into different levels. (c) We propose to employ the hierarchical time-series prediction model as an implicit proposal distribution of CMSMC method and formulate a unified tracking and prediction framework, which is able to handle occlusions and sensor failures.

## II. RELATED WORK

In this section, we provide a brief overview of the existing studies that are closely related to this work and point out the distinctiveness and novelty of the proposed approach.

### A. Tracking and Constrained State Estimation

In real-world tracking problems, there are commonly linear or nonlinear constraints on the estimated states. For instance, there are upper limits for the absolute value of acceleration and steering angle raised by dynamics feasibility restriction for vehicle tracking; the positions and velocities are also limited in proper ranges due to traffic rules. To enable incorporation of constraints, a constrained version of KF and its variants was proposed to handle linear or linearized systems [18], [19]. The constrained particle filter was also investigated to cope with nonlinear systems with complicated constraints through an acceptance/rejection procedure [20]. In [21], an additional optimization technique is employed to further improve the efficiency and robustness. The concept of mixture tracking was first proposed in [22] and applied to football player tracking in a sequence of video frames. In this work, we modify the original formulation of mixture tracking and generalize it to be adapted to general nonlinear discrete-time systems. Also, we add a constraint handling step in the mixture sequential Monte Carlo method to enhance tracking and prediction performance as well as to suggest potential constraint incorporation strategies. This step does not affect the theoretical convergence properties.

### B. Driver Behavior Recognition and Trajectory Prediction

Driver behavior recognition and vehicle trajectory prediction problems have been extensively investigated in literature. Widely used probabilistic models include Hidden Markov Model (HMM) [23]–[25], Gaussian Mixture Regression (GMR) [15], [26], Mixture Density Network (MDN) [27], Gaussian process (GP) [23], dynamic Bayesian network (DBN) [28], Rapidly-exploring Random Tree (RRT) [29], Variational Auto-Encoder (VAE) [30], [31], Generative Adversarial Network (GAN) [32]–[34] and multiple model approaches [35]. In this paper, we propose a hierarchical probabilistic model structure that can incorporate any of the above models for tracking and prediction. Moreover, instead of modeling each entity individually, we treat multiple interactive agents as a whole system and model the joint distribution of their future behaviors and motions.

## III. CONSTRAINED MIXTURE SEQUENTIAL MONTE CARLO (CMSMC)

In this section, we first present the theory of recursive constrained Bayesian state estimation with a mixture model representation. Since it is intractable and hard to obtain a closed-form estimated distribution, the constrained mixture sequential Monte Carlo approach is proposed to approximate the prior and posterior state distributions. The convergence analysis and practical implementation guides are provided.

### A. Recursive Constrained Mixture Bayesian State Estimation

Consider a general nonlinear discrete-time state space system with equality and (or) inequality constraints on the state which can be formulated as

$$\begin{aligned} \mathbf{x}_k &= q_{k-1}(\mathbf{x}_{k-1}, \mathbf{e}_{k-1}, \mathbf{v}_{k-1}), \\ \mathbf{z}_k &= h_k(\mathbf{x}_k, \mathbf{w}_k), \\ \mathbf{x}_k &\in \mathbb{S}_{\mathbf{x}_k}, \end{aligned} \quad (1)$$

where the subscript $k$ denotes the time step, $\mathbf{x}, \mathbf{e}, \mathbf{z}, \mathbf{v}, \mathbf{w}$ denotes the state vector, the exterior information, the measurement vector, the process noise and the measurement noise, respectively. Note that the random variable $\mathbf{e}$ is involved in our work since the state evolution can be affected by exterior factors, which rarely emerges in the canonical formulation. $\mathbb{S}_{\mathbf{x}_k}$ denotes the feasible state set satisfying all the constraints. $q(\cdot)$ represents the process model (a.k.a. system dynamics model) and $h(\cdot)$ represents the measurement model. The process model and measurement model can be time-invariant or time-variant and the noise value can be sampled from arbitrary distributions.

The recursive state-constrained Bayesian estimation consists of two steps:

**Step 1: Prior (Prediction) Update**:

$$f(\mathbf{x}_k|\mathbf{z}^{k-1}) = \int_{\mathbf{e}_{k-1}} \int_{\mathbf{x}_{k-1} \in \mathbb{S}_{\mathbf{x}_{k-1}}} [f(\mathbf{x}_k|\mathbf{x}_{k-1}, \mathbf{e}_{k-1}) \times f(\mathbf{x}_{k-1}, \mathbf{e}_{k-1}|\mathbf{z}^{k-1})] d\mathbf{e}_{k-1} d\mathbf{x}_{k-1}, \quad (2)$$

**Step 2: Measurement Update**:

$$f(\mathbf{x}_k|\mathbf{z}^k) = \frac{f(\mathbf{z}_k|\mathbf{x}_k) f(\mathbf{x}_k|\mathbf{z}^{k-1})}{\int_{\mathbf{x}_k \in \mathbb{S}_{\mathbf{x}_k}} f(\mathbf{z}_k|\mathbf{x}_k) f(\mathbf{x}_k|\mathbf{z}^{k-1}) d\mathbf{x}_k}, \quad (3)$$

where $f(\cdot)$ represents the probability density function and $\mathbf{z}^k = (\mathbf{z}_1, \cdots, \mathbf{z}_k)$ represents the measurement up to time step $k$. The initial state distribution is set to be $f(\mathbf{x}_0|\mathbf{z}_0)$ which is adaptive to the initial measurement.

The above formulation works well for estimating unimodal distributions which is widely used in single object tracking. However, it does not perform well in multi-modal distribution estimation as well as in multi-object tracking problems since the estimated distribution tends to degenerate to be unimodal along time (e.g. due to particle weight degeneracy in SMC). Therefore, the mixture model formulation is utilized to maintain multiple modalities, which requires no distribution parameterization assumptions. The posterior state distribution can be written as

$$f(\mathbf{x}_k|\mathbf{z}^k) = \sum_{m=1}^{M} \pi_{m,k} f_m(\mathbf{x}_k|\mathbf{z}^k), \quad (4)$$

where $M$ denotes the component number of the mixture model, $\pi_{m,k}$ denotes the $m$-th component weight at time step $k$ and $\sum_{m=1}^{M} \pi_{m,k} = 1$. Assuming that the posterior state distributions for each mixture component at time step $k-1$, i.e. $f_m(\mathbf{x}_{k-1}|\mathbf{z}^{k-1})$ has been obtained from the last measurement update and the exterior information $\mathbf{e}_{k-1}$ is independent from $\mathbf{x}_{k-1}$, we can calculate the new prior state distribution by

$$f(\mathbf{x}_k|\mathbf{z}^{k-1}) = \sum_{m=1}^{M} \pi_{m,k-1} \int_{\mathbf{e}_{k-1}} \int_{\mathbf{x}_{k-1}} [f_m(\mathbf{x}_k|\mathbf{x}_{k-1}, \mathbf{e}_{k-1}) \times f_m(\mathbf{x}_{k-1}|\mathbf{z}^{k-1}) f_m(\mathbf{e}_{k-1}|\mathbf{z}^{k-1})] d\mathbf{x}_{k-1} d\mathbf{e}_{k-1}. \quad (5)$$

When a new measurement is taken in, the prior state distribution is substituted into (3), which leads to

$$f(\mathbf{x}_k|\mathbf{z}^k) = \frac{\sum_{m=1}^{M} \pi_{m,k-1} f_m(\mathbf{z}_k|\mathbf{x}_k) f_m(\mathbf{x}_k|\mathbf{z}^{k-1})}{\sum_{n=1}^{M} \pi_{n,k-1} \int_{\mathbf{x}_k \in \mathbb{S}_{\mathbf{x}_k}} f_n(\mathbf{z}_k|\mathbf{x}_k) f_n(\mathbf{x}_k|\mathbf{z}^{k-1}) d\mathbf{x}_k}. \quad (6)$$

The new posterior distribution and mixture weight for the $m$-th component can be obtained through following equations:

$$f_m(\mathbf{x}_k|\mathbf{z}^k) = \frac{f_m(\mathbf{z}_k|\mathbf{x}_k) f_m(\mathbf{x}_k|\mathbf{z}^{k-1})}{\int_{\mathbf{x}_k \in \mathbb{S}_{\mathbf{x}_k}} f_m(\mathbf{z}_k|\mathbf{x}_k) f_m(\mathbf{x}_k|\mathbf{z}^{k-1}) d\mathbf{x}_k}, \quad (7)$$

$$\pi_{m,k} = \frac{\pi_{m,k-1} \int_{\mathbf{x}_k \in \mathbb{S}_{\mathbf{x}_k}} f_m(\mathbf{z}_k|\mathbf{x}_k) f_m(\mathbf{x}_k|\mathbf{z}^{k-1}) d\mathbf{x}_k}{\sum_{n=1}^{M} \pi_{n,k-1} \int_{\mathbf{x}_k \in \mathbb{S}_{\mathbf{x}_k}} f_n(\mathbf{z}_k|\mathbf{x}_k) f_n(\mathbf{x}_k|\mathbf{z}^{k-1}) d\mathbf{x}_k}$$

$$= \pi_{m,k-1} f_m(\mathbf{z}_k|\mathbf{z}^{k-1}) / \sum_{n=1}^{M} \pi_{n,k-1} f_n(\mathbf{z}_k|\mathbf{z}^{k-1}). \quad (8)$$

The above recursive process can be applied to each individual component which only interacts with others by the adaptive adjustment of component weights in each iteration.

### B. CMSMC Approximation

In order to approximate the constrained mixture state estimation recursion, we propose a constrained mixture sequential Monte Carlo approach which can adopt arbitrary state evolution models as implicit proposal distribution. The CMSMC approximation is represented by six sets of variables: state vector $\mathbf{x}$, particle normalized weight $w$, particle unnormalized weight $\bar{w}$, the component identity $c$ that indicates which component the particle pertains to, and the feasibility indicator $\mathbb{I}$ which reveals whether the state vector of this particle is inside the feasible region. We set $\mathbb{I} = 1$ for feasible particles while $\mathbb{I} = 0$ for unfeasible ones. A self-contained particle state formulation is defined as

$$\mathbf{p}_k^{(i)} = [\mathbf{x}_k^{(i)} \ w_k^{(i)} \ \bar{w}_k^{(i)} \ c_k^{(i)} \ \pi_k^{(i)} \ \mathbb{I}_k^{(i)}], \quad (9)$$

where the subscript $k$ denotes the time step and the superscript $(i)$ denotes the particle identity.

Rather than only estimate the state vector at the current time step, the proposed CMSMC method is able to estimate the whole state profile, which gives

$$\hat{f}(\mathbf{x}^{k(i)}|\mathbf{z}^{k-1}) = \sum_{m=1}^{M} f_m(\mathbf{x}^{k(i)}|\mathbf{x}^{k-1(i)}, \mathbf{e}^{k-1}, \mathbf{z}^{k-1}) \times f_m(\mathbf{x}^{k-1(i)}, \mathbf{e}^{k-1}|\mathbf{z}^{k-1}), \quad (10)$$

where $\mathbf{x}^{k(i)} = (\mathbf{x}_1^{(i)}, \cdots, \mathbf{x}_k^{(i)})$ represents the whole trajectory of the $i$-th particle. Since the algorithm complexity will increase much if a marginalization process is implemented to obtain $\hat{f}(\mathbf{x}_k|\mathbf{z}^k)$ from $\hat{f}(\mathbf{x}^k|\mathbf{z}^k)$, a widely used simplification [8] is employed as

$$f(\mathbf{x}_k|\mathbf{z}^k) \approx \hat{f}(\mathbf{x}_k|\mathbf{z}^k) = \sum_{m=1}^{M} \pi_{m,k} \sum_{i \in \mathcal{C}_m} w_k^{(i)} \delta(\mathbf{x}_k^{(i)} - \mathbf{x}_k),$$

$$\sum_{m=1}^{M} \pi_{m,k} = 1, \quad \sum_{i \in \mathcal{C}_m} w_k^{(i)} = 1, \ m = 1, \ldots, M, \quad (11)$$

where $\delta(\cdot)$ denotes the Dirac delta function and $\mathcal{C}_m$ denotes the particle identity set corresponding to the $m$-th component. The details of the CMSMC procedure are introduced below, where we denote $N_p$ as the total amount of particles.

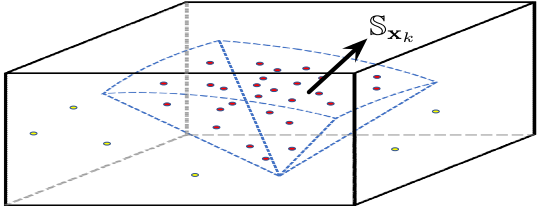

Fig. 1. An illustrative diagram of incorporating state constraints. (The whole state space is represented by the cube and the feasible region is represented by a polytope which is restricted by four linear inequality constraints and one nonlinear inequality constraint. The red points and yellow points signify the feasible particles and unfeasible ones which may be rejected or assigned a zero weight, respectively.)

*1) Initialization:* The initial particles are randomly sampled from the feasible regions of initial state distribution $f(\mathbf{x}_0|\mathbf{z}_0)$ with equal weight $1/N_p$, which automatically guarantees that the state constraints are satisfied.

*2) Prior (Prediction) Update & Measurement Update:* The sequential importance sampling (SIS) technique [8] is employed in this step. We incorporate a proposal distribution $f_m^p(\mathbf{x}_k|\mathbf{x}_{k-1}, \mathbf{e}_{k-1}, \mathbf{z}_k)$ for each component from which the new particles are sampled and obtain that

$$f(\mathbf{x}_k|\mathbf{z}^{k-1}) = \sum_{m=1}^{M} \pi_{m,k-1} \int_{\mathbf{e}_{k-1}} \int_{\mathbf{x}_{k-1} \in \mathbb{S}_{\mathbf{x}_{k-1}}} [f_m^p(\mathbf{x}_k|\mathbf{x}_{k-1}, \mathbf{e}_{k-1}, \mathbf{z}_k) \frac{f_m(\mathbf{x}_k|\mathbf{x}_{k-1}, \mathbf{e}_{k-1})}{f_m^p(\mathbf{x}_k|\mathbf{x}_{k-1}, \mathbf{e}_{k-1}, \mathbf{z}_k)} f_m(\mathbf{x}_{k-1}, \mathbf{e}_{k-1}|\mathbf{z}^{k-1})] d\mathbf{e}_{k-1} d\mathbf{x}_{k-1}. \quad (12)$$

Here we provide two strategies of incorporating constraints:

(i) For the $m$-th mixture component, sample a new state vector $\mathbf{x}_k$ for each existing particle from the proposal distribution $f_m^p(\mathbf{x}_k|\mathbf{x}_{k-1}, \mathbf{e}_{k-1}, \mathbf{z}_k)$ only once and check whether $\mathbf{x}_k \in \mathbb{S}_{\mathbf{x}_k}$ is satisfied. Set $\mathbb{I}_k^{(i)}$ to be unity if $\mathbf{x}_k \in \mathbb{S}_{\mathbf{x}_k}$; otherwise, set $\mathbb{I}_k^{(i)}$ to be zero. Then calculate the new weights for the particles by

$$\bar{w}_k^{(i)} = \mathbb{I}_k^{(i)} \cdot w_{k-1}^{(i)} \frac{f_m(\mathbf{x}_k^{(i)}|\mathbf{x}_{k-1}^{(i)}, \mathbf{e}_{k-1}) f_m(\mathbf{z}_k|\mathbf{x}_k^{(i)})}{f_m^p(\mathbf{x}_k^{(i)}|\mathbf{x}_{k-1}^{(i)}, \mathbf{e}_{k-1}, \mathbf{z}_k)}, \quad (13)$$

$$w_k^{(i)} = \bar{w}_k^{(i)} / \sum_{j \in \mathcal{C}_m} \bar{w}_k^{(j)}. \quad (14)$$

The posterior distribution of the $m$-th component can be properly approximated by the new particle set.

(ii) An alternative strategy is similar to (i) except that the new state vector $\mathbf{x}_k$ needs re-sampling until $\mathbf{x}_k \in \mathbb{S}_{\mathbf{x}_k}$ is satisfied. The corresponding particle weight is updated by (15) instead of (13)

$$\bar{w}_k^{(i)} = w_{k-1}^{(i)} \frac{f_m(\mathbf{x}_k^{(i)}|\mathbf{x}_{k-1}^{(i)}, \mathbf{e}_{k-1}) f_m(\mathbf{z}_k|\mathbf{x}_k^{(i)})}{f_m^p(\mathbf{x}_k^{(i)}|\mathbf{x}_{k-1}^{(i)}, \mathbf{e}_{k-1}, \mathbf{z}_k)}. \quad (15)$$

The advantages and weaknesses of each strategy are discussed in Section III-D.

After obtaining the new particle weights, the component weights can be maintained accordingly by

$$\pi_{m,k} = \frac{\pi_{m,k-1} \int_{\mathbf{x}_k \in \mathbb{S}_{\mathbf{x}_k}} f_m(\mathbf{z}_k|\mathbf{x}_k) f_m(\mathbf{x}_k|\mathbf{z}^{k-1}) d\mathbf{x}_k}{\sum_{n=1}^{M} \pi_{n,k-1} \int_{\mathbf{x}_k \in \mathbb{S}_{\mathbf{x}_k}} f_m(\mathbf{z}_k|\mathbf{x}_k) f_m(\mathbf{x}_k|\mathbf{z}^{k-1}) d\mathbf{x}_k}$$

$$\approx \pi_{m,k-1} \sum_{i \in \mathcal{C}_m} \bar{w}_k^{(i)} / \sum_{n=1}^{M} \pi_{n,k-1} \sum_{j \in \mathcal{C}_n} \bar{w}_k^{(j)}. \quad (16)$$

There can be a particle re-sampling step for each component like canonical sequential Monte Carlo methods to avoid weight degeneracy if necessary.

*3) Reclustering:* Due to the long-term evolution of particles, the particle belonging to one component may become spatially close to another component, which requires a re-clustering step. Without loss of generality, in this work we employ the $k$-medoids approach where $k$ equals the number of components. This does not change the estimated posterior distribution thus not affect the convergence analysis in Section III-C.

*4) Mixture Component Weight Update:* Since particles may transfer among different components, the component weights need to be updated to maintain the same distribution by

$$\hat{f}(\mathbf{x}_k|\mathbf{z}^k) = \sum_{m=1}^{M} \pi'_{m,k} \sum_{i \in \mathcal{C}'_m} w'^{(i)}_k \delta(\mathbf{x}_k^{(i)} - \mathbf{x}_k),$$

$$\pi'_{m,k} = \sum_{i \in \mathcal{C}'_m} \pi_{c_k^{(i)},k} w_k^{(i)}, \quad w'^{(i)}_k = (\pi_{c_k^{(i)},k} / \pi'_{c_k^{(i)},k}) w_k^{(i)}. \quad (17)$$

### C. Convergence Analysis

Many research efforts have been devoted to investigate the theoretical convergence properties of canonical sequential Monte Carlo methods such as [36], [37]. This subsection provides a concise convergence analysis of the proposed CMSMC method based on the following propositions for canonical SMC adapted from [37] where the exhaustive proofs can be found.

*Proposition 1:* If the state transition distribution $f(\mathbf{x}_k|\mathbf{x}_{k-1})$ is continuous and for $\forall k$, $0 < f(\mathbf{z}_k|\mathbf{x}_k) \leq C(k, \mathbf{z}_k) < \infty$ is satisfied, then for $\forall k$ and $\forall \mathbf{z}^k$, $||\hat{f}(\mathbf{x}_k|\mathbf{z}^k) - f(\mathbf{x}_k|\mathbf{z}^k)||_1 \to 0$ as the particle number $N_p \to \infty$.

*Proposition 2:* Under the conditions in Proposition 1, the approximated distribution by particles $\hat{f}(\mathbf{x}_k|\mathbf{z}^k)$ converges to the true posterior distribution at the rate of $1/\sqrt{N_p}$.

Due to the decomposability of mixture components in the CMSMC formulation, the convergence property can be evaluated at component level. The state transition distribution can be calculated as

$$f(\mathbf{x}_k|\mathbf{x}_{k-1}) = \sum_{m=1}^{M} \pi_{m,k-1} \int_{\mathbf{e}_{k-1}} f_m(\mathbf{x}_k|\mathbf{x}_{k-1}, \mathbf{e}_{k-1}), \quad (18)$$

which is continuous and the likelihood value $f_m(\mathbf{z}_k|\mathbf{x}_k)$ is bounded. Moreover, the reclustering process does not essentially modify the particle representation. Therefore, the convergence property in Proposition 1 also applies.

*D. Practical Implementation Guides*

*1) Constraint Incorporation:* The two constraint incorporation strategies have advantages on different aspects. Therefore, the choice should be made according to emphasis on the performance in a particular problem. The first strategy can guarantee the feasibility of the nonzero-weighted particles as well as keep the least computational cost since there is no multiple sampling process for a particular particle. This works well if the proposal distribution is properly chosen and the constraints are not hard to satisfy. Otherwise, there will be more and more rejected particles along time which results in a significant reduction of particle amount and low estimation quality even divergence. Under such situations, the second strategy which maintains a constant size of the particle set is the better choice despite larger computational cost.

There is another intuitive strategy which only samples the new particles once and pushes the unfeasible ones to the boundary of the feasible region. However, this is very hard to implement, especially when the boundary is highly nonlinear. Moreover, it may result in a high density around the boundary which leads to a large deviation to the true distribution.

*2) Proposal Distribution:* In order to obtain a approximated posterior distribution, the optimal proposal distribution is

$$f_m^p(\mathbf{x}_k|\mathbf{x}_{k-1}^{(i)}, \mathbf{e}_{k-1}, \mathbf{z}_k) = f_m(\mathbf{x}_k|\mathbf{x}_{k-1}^{(i)}, \mathbf{e}_{k-1}, \mathbf{z}_k), \quad (19)$$

which has no influence on the particle weight variance. However, it is usually difficult to properly sample from this distribution and the weight update process brings much computational cost. A widely used alternative is

$$f_m^p(\mathbf{x}_k|\mathbf{x}_{k-1}^{(i)}, \mathbf{e}_{k-1}, \mathbf{z}_k) = f_m(\mathbf{x}_k|\mathbf{x}_{k-1}^{(i)}, \mathbf{e}_{k-1}), \quad (20)$$

which is easy to sample from and can significantly simplify the weight update equation (13) and (15) into

$$\bar{w}_k^{(i)} = \mathbb{I}_k^{(i)} \cdot w_{k-1}^{(i)} f_m(\mathbf{z}_k|\mathbf{x}_k^{(i)}), \quad (21)$$
$$\bar{w}_k^{(i)} = w_{k-1}^{(i)} f_m(\mathbf{z}_k|\mathbf{x}_k^{(i)}). \quad (22)$$

In this work, we employed this proposal distribution in all the experiments.

*3) Resampling Strategy:* In order to avoid weight degeneracy, the particles need to be resampled when necessary. A straightforward way is to resample at each iteration. However, this is only suitable for offline implementation since it may ruin the real-time capability of the algorithm due to the large computational cost. A better choice is to use the effective number criterion [8] where

$$N_{\text{eff},k} = \frac{N_p}{1 + \frac{\Sigma(w_k^{(i)})}{[\mu(w_k^{(i)})]^2}} \approx \frac{1}{\sum_{i=1}^{N_p}(w_k^{(i)})^2}. \quad (23)$$

If $N_{\text{eff},k}$ is less than a proper threshold $N_{th}$, then the resampling process applies. A comparison study of the resampling algorithms is provided in [38]. In this work, we utilized systematic resampling in all the experiments.

*4) Real-Time Capability:* The real-time performance of proposed CMSMC approach is mainly dependent on the following factors: i) common: particle amount, resampling frequency and constraint incorporation strategy; ii) problem-specific: target number, model complexity and proposal distribution sampling efforts. It is natural that more particles lead to a better approximation of distributions. However, in many cases it is likely that when the particle amount is large enough, adding particles will increase computational cost but bring little performance improvement. Therefore, a tradeoff on particle amount should be made by choosing the best setup through multiple experiments. Another consideration is to dynamically adjust particle number such as reduce particles after convergence and add particles when they tend to diverge or the tracking accuracy tends to decrease.

*5) Divergence Alert:* In real-time applications, it is crucial to set up a divergence alert mechanism to decide whether to adjust particle amount or even reinitialize the algorithm. There are two efficient divergence indicators: the effective number $N_{\text{eff}}$ and raw likelihood values of particles. If they are less than properly chosen thresholds, the divergence alert will be activated.

## IV. HIERARCHICAL TIME-SERIES PREDICTION MODEL (HTSPM)

In this section, we propose a hierarchical prediction model for time-series problems which consists of two modules: recognition module and evolution module. The recognition module aims at solving a probabilistic classification problem while the evolution module aims at propagating the current state to the future. Mathematically, the proposed model is used to approximate the state transition distribution

$$\hat{f}(\mathbf{x}_k|\mathbf{x}_{k-1}, \mathbf{e}_{k-1}) = \sum_B \int_{\mathbf{u}_{k-1}} \int_{\mathbf{z}^k} \hat{f}(\mathbf{x}_k|\mathbf{x}_{k-1}, \mathbf{u}_{k-1})$$
$$\times \hat{f}(\mathbf{u}_{k-1}|\mathbf{x}_{k-1}, \mathbf{e}_{k-1}, B_{k-1})$$
$$\hat{f}(B_{k-1}|\mathbf{z}^k) d\mathbf{u}_{k-1} d\mathbf{z}^k, \quad (24)$$

where $B$ denotes behavior class set elements. The model details and advantages are presented in Section IV-A,B and potential applications are discussed in Section IV-C.

*A. Recognition Module*

The recognition module aims at obtaining the posterior probabilities of classes $\hat{f}(B_{k-1}|\mathbf{z}^k)$ where $B_{k-1}$ is a discrete random variable representing the class at time step $k-1$. In many applications, it is reasonable to make an assumption that $\hat{f}(B_{k-1}|\mathbf{z}^k) \approx \hat{f}(B_{k-1}|\mathbf{z}^{k-T:k})$ where $\mathbf{z}^{k-T:k} = (\mathbf{z}_{k-T}, \ldots, \mathbf{z}_k)$ and $T$ is a properly chosen period length according to specific problem setups since the most recent several observations have far more significance than the past ones when deciding the current behavior class probabilities. There is no limitation on the specific architecture of recognition model provided it provides posterior class probabilities. Although there exist many widely used probabilistic classifiers that can be directly employed in this module, we propose a

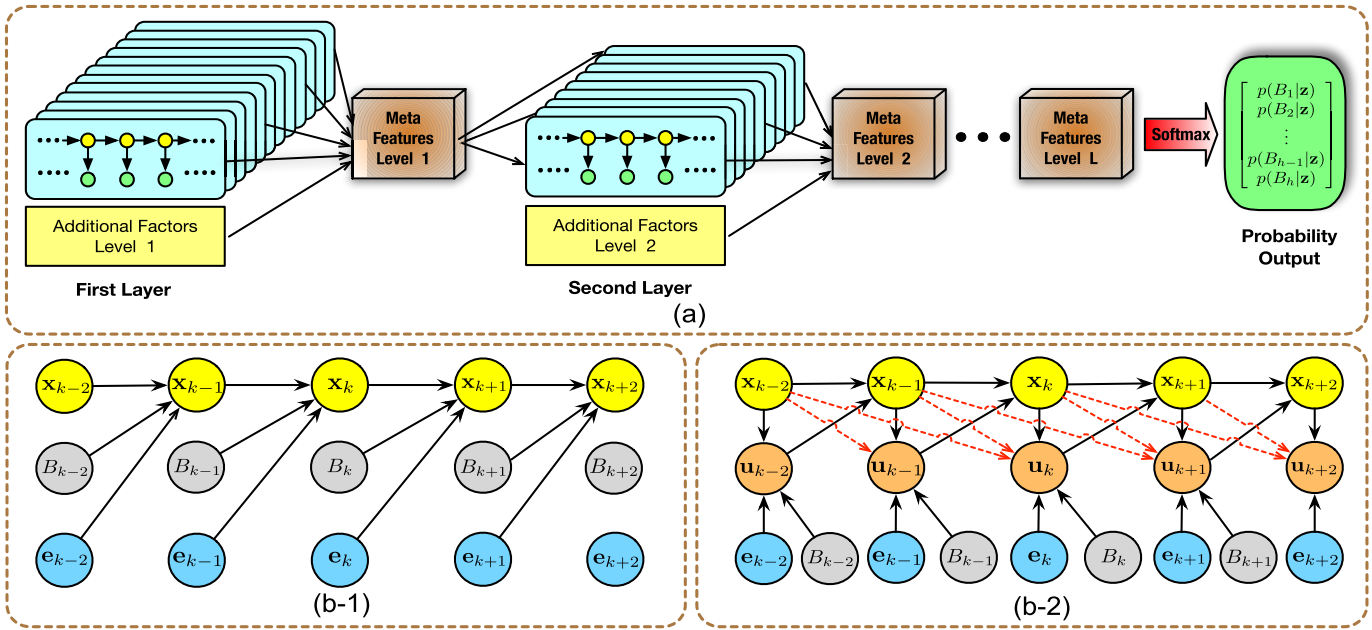

Fig. 2. The diagram of hierarchical time-series prediction model. (a) Recognition module: Deep Hidden Markov Model (DHMM); (b-1) Evolution module: the exterior information affects the state directly; (b-2) Evolution module: the exterior information and behavior pattern affect the state indirectly through an action (input) term. The black solid arrows represent first-order Markov assumption and the red dashed ones imply higher order assumptions.

*Deep Hidden Markov Model (DHMM)* which has advantages over existing ones in multiple aspects.

The DHMM has a multi-layer architecture, which is shown in Fig. 2. Each layer is composed of a group of canonical Hidden Markov Models (HMM). In general, the $l$-th layer HMM take in the outputs of the $(l-1)$-th layer and extract meta features which are used as observations for the $(l+1)$-th layer HMM. For instance, the first-layer HMM take in the lowest-level observations such as raw sensor measurement of state as well as exterior information and extract meta features which are utilized as observations of the second-layer HMM. Since there may be additional factors of different levels affecting the recognition results, they are incorporated into the meta features at proper levels. This information extraction and message passing mechanism applies to both training and recognition phases. After extracting the meta features of the last layer, there is a *Softmax* layer calculating the posterior probabilities. In this work, we use the observation log-likelihood of each layer HMM as the meta features. An alternative for meta features are the class indices with highest likelihood which results in degeneracy into deterministic recognition except the output layer, which is suitable when large distinctions exist among different classes. This can also reduce computational efforts. In this paper, we used the former in all the experiments.

*1) Training Phase:* The whole training trajectories are divided into proper segments at different levels and properly labeled. The DHMM is trained from the first layer to the last layer successively. More specifically, assuming that there are $h_l$ classes in the $l$-th layer, then $h_l$ HMM are trained with log-likelihood sequences obtained from the $(l-1)$-th layer output using the Baum-Welch algorithm (a.k.a Forward-Backward algorithm) [39]. In order to choose the best hidden state number, we can use the Bayesian information criterion (BIC) as a performance indicator by pre-fitting a Gaussian mixture distribution to the training feature sequences.

*2) Recognition Phase:* Given a new observation sequence, we also use a bottom-up procedure similar to the training phase to obtain the posterior class probabilities. The implementation details are summarized in the first half of **Algorithm 1**, where $T_l$ is the period length for the $l$-th layer which needs to be tuned properly for different problems and application scenarios. If it is too large, the extracted meta feature sequence will have a short length, which passes less observations to future layers; if it is too small, the quality of current-layer output will decrease. In practice, a possible issue is the observation sequence likelihood obtained by different class HMM may be in different number scales, which will lead to a class with an absolute dominant probability ($\approx 1$) at all time. Therefore, it is recommended to incorporate a set of calibration parameters $\alpha_h$, $h = 1, 2, \ldots, h_l$ to make equal probabilities for each class at the initial time step and keep the same parameter values afterwards.

We select several popular probabilistic classifiers as baseline models and make comparisons in the numerical case study.

*a) Standard hidden markov model (HMM):* Instead of partitioning the high-level behaviors into multiple stages like DHMM, HMM classifier treats the entire trajectory as a whole. In the training phase, the whole sequences are utilized to train the HMM by Baum-Welch algorithm. In the recognition phase, a segment of historical information are fed to HMM to obtain the likelihood and normalized into probabilities.

*b) Gaussian discriminant analysis (GDA):* The essence of GDA is to obtain a linear decision surface for Linear Discriminant Analysis (LDA) or a quadratic decision surface for Quadratic Discriminant Analysis (QDA) through proper transformations of raw features which can distinguish among

**Algorithm 1** HTSPM Prediction Algorithm

**Input:**
  1. The number of layers $L$ and well-trained HMM of all layers HMM-$l$-$h$ ($l = 1, \ldots, L$; $h = 1, \ldots, h_l$);
  2. The test raw observation sequence;
  3. The last step particle hypotheses $\{\mathbf{x}_{k-1}^{(j)}, j = 1, \ldots, N_p\}$.

**Output:**
  The current particle hypotheses $\{\mathbf{x}_k^{(j)}, j = 1, \ldots, N_p\}$.

1: Recognition phase:
2: $obs \leftarrow$ raw observation sequence;
3: $len \leftarrow \text{Length}(obs)$;
4: **for** $l = 1, 2, \ldots, L$ **do**
5:   **for** $h = 1, 2, \ldots, h_l$ **do**
6:     **for** $i = 1, 2, \ldots, len - T_l$ **do**
7:       $\mathcal{L}_{lh}$.append(Likelihood(HMM-$l$-$h$, $obs[i : i + T_l]$));
8:     **end for**
9:   **end for**
10:   $\mathcal{L}_l \leftarrow$ concatenating $\mathcal{L}_{lh}$;
11:   $obs \leftarrow \mathcal{L}_l$; $len \leftarrow \text{Length}(obs)$;
12: **end for**
13: $probability \leftarrow \text{Softmax}(obs)$;
14: Evolution phase:
15: **for** $j = 1, 2, \ldots, N_p$ **do**
16:   Sample $B_{k-1}^{(j)}$ from $probability$;
17:   Sample $\mathbf{u}_{k-1}^{(j)}$ from $\hat{f}(\mathbf{u}_{k-1}|\mathbf{x}_{k-1}, \mathbf{e}_{k-1}, B_{k-1}^{(j)})$;
18:   **if** $\mathbf{u}_{k-1}^{(j)}$ not feasible **then**
19:     Resample from proposal distribution;
20:   **else**
21:     $\mathbf{x}_k^{(j)} \leftarrow \mathcal{F}(\mathbf{x}_{k-1}^{(j)}, \mathbf{u}_{k-1}^{(j)})$;
22:   **end if**
23: **end for**
24: **return** $\{\mathbf{x}_k^{(j)}, j = 1, \ldots, N_p\}$

categories and perform classification in the transformed space according to some distance metric such as Euclidean distance. While LDA assumes that all the classes share the same covariance matrix, QDA fits a particular covariance matrix for each class. The feature matrix consists of historical observation information and the labels are behavior indices.

    *c) Gaussian naive bayes (GNB):* NB is a typical probabilistic classifier which employs Bayes' theorem and assumes that features are mutually independent [40]. In this work, we assume that the feature likelihood to be Gaussian distribution, which establishes a GNB.

The advantages of the proposed DHMM over above baseline models are four folds: *1)* Compared with the models which only take raw observations as input features, our model is able to extract multi-level features and robust to measurement noise and sensor failures; *2)* Compared with other deep models such as deep neural networks, our model requires a significantly less amount of training data and computational cost as well as maintains interpretability from a probability perspective; *3)* Thanks to the layered representation and decomposability between layers, DHMM has potential knowledge transferability among similar tasks. It will reduce much training efforts if we can utilize several parameters directly or finetune from well-trained models for new tasks; *4)* The training and recognition processes can be parallelized since the learning and inference of HMM are independent within a layer.

### B. Evolution Module

The evolution module is designed to obtain the conditional state transition distribution given a certain behavior pattern $\hat{f}(\mathbf{x}_k|\mathbf{x}_{k-1}, \mathbf{e}_{k-1}, B_{k-1})$ which is demonstrated in Fig. 2(b-1) where the exterior information has effects on the state directly. When the exterior information and behavior pattern affect the state indirectly through an action term, the conditional distribution can be further extended to $\hat{f}(\mathbf{x}_k|\mathbf{x}_{k-1}, \mathbf{u}_{k-1})\hat{f}(\mathbf{u}_{k-1}|\mathbf{x}_{k-1}, \mathbf{e}_{k-1}, B_{k-1})$ which is presented in Fig. 2(b-2). The detailed procedures of evolution phase can be found in the second half of **Algorithm 1**. In this work, we demonstrate three learning-based state evolution models and compare their performance in the experiments.

*1. Conditional Gaussian Mixture Regression (CGMR):* The driver behavioral model proposed in the authors' previous work [26] is adapted and generalized as CGMR which is based on a Gaussian mixture model (GMM). The conditional Gaussian mixture distribution is a linear combination of multiple Gaussians with the form $f(\zeta^i) = \sum_{g=1}^{N} \pi_g^i \mathcal{N}(\zeta^i|\mu_g^i, \Sigma_g^i)$ where $i$ is the behavior class index, $\sum_{g=1}^{N} \pi_g^i = 1$, $\mu_g^i$ and $\Sigma_g^i$ are the mean and covariance of the $g$-th Gaussian distribution, and $\zeta^i$ is the training dataset for the $i$-th behavior. In each training sample, the input and output are stacked into a column vector which is denoted as $\zeta^i = [\ \mathcal{I}^i\ |\ \mathcal{O}^i\ ]^T$, where $\mathcal{I}^i$ denotes the conditional variables and $\mathcal{O}^i$ denotes the predicted variables. The dimensions of the two variables are arbitrary. For instance, in Fig. 2(b-2) the $\mathbf{e}_{k-1}$, $B_{k-1}$ and $\mathbf{x}_{k-1}$ can be treated as conditional variables while $\mathbf{u}_{k-1}$ and $\mathbf{x}_k$ can be treated as the corresponding predicted variables. The training and prediction method are identical to [26].

*2. Conditional Probabilistic Multi-Layer Perceptrons (CP-MLP):* MLP is a subclass of deep neural network which consists of a directed acyclic feed-forward architecture [41]. The network input is historical state information and the output are the actions in a certain length of time horizon. To improve generality, an $L$2-regularization term is added in the loss function and dropout layers are incorporated. However, since canonical MLP is a deterministic model, we add a noise term sampled from normal distribution to the network input to incorporate uncertainty during both training and test process. We train a P-MLP for each behavior class, which establishes a set of CP-MLP.

*3. Conditional Probabilistic Long Short-Term Memory (CP-LSTM):* LSTM is a widely used variant of recurrent neural network (RNN) which is suitable for time-series data modeling and can effectively avoid gradient explosion and vanishing issues [42]. The network takes in a sequence of historical state and gives out a sequence of future actions. Similar to CP-MLP, a noise term is also appended to the input features to involve uncertainty. The CP-LSTM has a similar architecture to CP-MLP except that the first hidden layer is replaced with a LSTM layer.

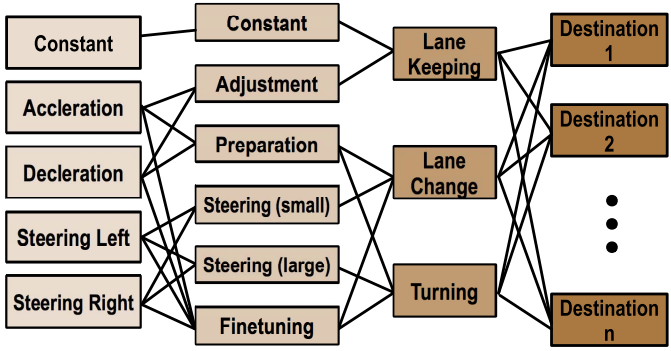

Fig. 3. The hierarchical representation of driver behaviors. In daily driving situations, there are three common behaviors that can be arranged to get to any destination accordingly: lane keeping, lane change and turning. These behaviors can also be decomposed to more primary actions such as speed adjustment and steering which also have composing elements. Each behavior level corresponds to a layer and each behavior class at a certain level corresponds to an HMM.

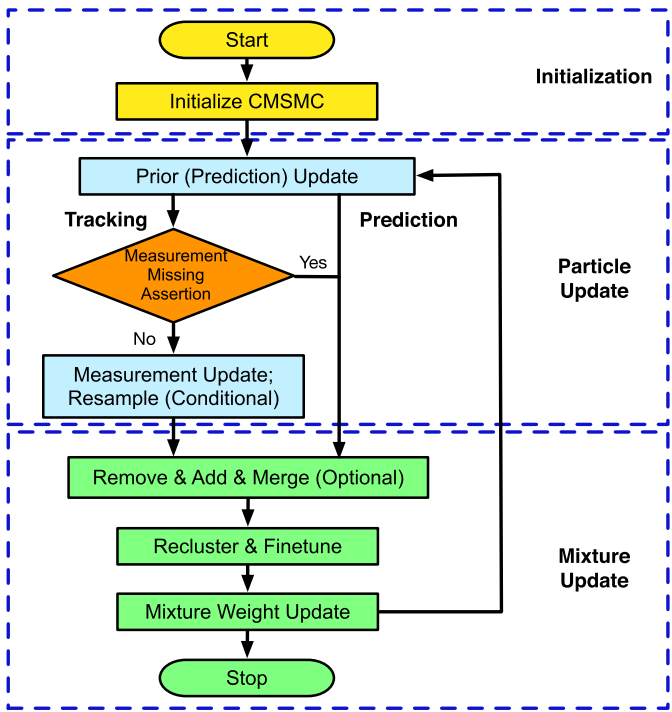

Fig. 4. The diagram of CMSMC-based tracking and prediction framework.

### C. Application Scopes

The proposed hierarchical time-series model is especially suitable for recognition and prediction of complicated events consisted of multi-level sub-stages, such as driver behaviors and human activities; or classification problems in which the high-level category has several sub-classes, such as in natural language processing. An illustration of exemplar application scenarios specifically for driver behaviors can be found in Fig. 3. In this paper, the application to vehicle tracking and prediction is validated and discussed in the case study.

### V. GENERIC TRACKING AND PREDICTION FRAMEWORK

In this section, we propose a generic tracking and prediction framework based on the constrained mixture sequential Monte Carlo approach, whose flow diagram is illustrated in Fig. 4.

**Algorithm 2** CMSMC-Based Tracking and Prediction
**Input:**
    1. Function Mode ($FM$): tracking (0) or prediction (1);
    2. Mixture Update Mode ($MUM$): fixed component number (0) or adaptive component number (1);
    3. Initial component number $M_0$;
    4. Initial particle amount for each component $n_c$;
       the total particle amount is $N_p = M_0 \, n_c$ accordingly;
1:  $StopFlag \leftarrow 0; \quad k \leftarrow 1;$
2:  Initialization: draw initial particles $\{\mathbf{x}_0^{(i)} : i = 1, \ldots, N_p\}$ according to a known $f(\mathbf{x}_0|\mathbf{z}_0)$ with equal weights;
3:  **while** $StopFlag = 0$ **do**
4:     Prior Update: sample $\mathbf{x}_k^{(i)}$ from the proposal distribution $\mathbf{x}_k^{(i)} \sim f_m^p(\mathbf{x}_k|\mathbf{x}_{k-1}^{(i-1)}, \mathbf{e}_{k-1}, \mathbf{z}_k)$ using one of the constraint incorporation strategies (use Algorithm 1);
5:     **if** $FM = 0$ **then**
6:         Measurement Update: calculate the unnormalized particle weights by (13) or (15) and normalized weights by (14); If $N_{\text{eff},k} < N_{th}$, resample by the systematic resampling algorithm or any other proper strategies;
7:     **end if**
8:     **if** $MUM = 1$ **then**
9:         Adjust the component number adaptively as illustrated in Section III and obtain the new component number $M'$;  $M \leftarrow M';$
10:    **end if**
11:    Recluster the particles and calculate the new component weights by (17);  $k \leftarrow k + 1;$
12:  **end while**

The framework has a closed-loop structure which falls into three stages: initialization, particle update and mixture update. A summarized implementation procedure of the framework can be found in **Algorithm 2**.

The framework has two function modes: tracking mode and prediction mode. In each iteration of the tracking mode, there is a "measurement missing assertion" step through setting a proper distance threshold to check whether the new measurement of tracking targets are lost due to complete occlusion or sensing failure. If so, the current step is treated as a prediction problem thus without measurement update.

In the real-world applications, the number of tracking targets may fluctuate along time due to object emergence and disappearance as well as merging and splitting. Therefore, an adaptive adjustment mechanism is required so that the difference between component number and true target quantity is minimized. Therefore, a "Remove & Add & Merge" step is employed to adaptively adjust the component number, which is introduced in detail below.

(i) *Remove*: the components are removed if the corresponding weights are less than $\pi_{th}$ or the mean point is outside the observation area;

(ii) *Add*: the component number will increase by one if the amount of particles assigned to a certain measurement in the last iteration is less than $N_{mth}$ which we treat as a new target emergence. New particles are drawn around the new target;

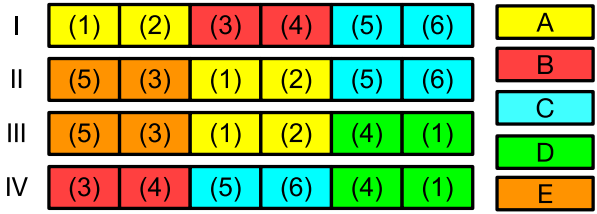

Fig. 5. The diagram of hierarchical behavior representation in the numerical case. There are four high-level behaviors I, II, III and IV which are composed of three of the five stages A, B, C, D and E while each of these stages can also be separated into two of the six sub-stages (1)-(6). The order of stages and sub-stages are fixed in each high-level behavior.

TABLE I
THE FUNCTION $g(k)$ FOR EACH SUB-STAGE

| Sub-stages | $g(k)$ | Sub-stages | $g(k)$ | Sub-stages | $g(k)$ |
|---|---|---|---|---|---|
| (1) | 1.5 | (2) | $1.5\cos(0.1k)$ | (3) | -0.75 |
| (4) | $3\sin(0.1k)$ | (5) | -3 | (6) | 3 |

(iii) *Merge*: the two components are merged when their distance is less than $d_{th}$. We employ the distance metric proposed in [43] for component $m$ and $n$

$$dist(m,n) = \frac{\int [\hat{f}_m(\mathbf{x}) - \hat{f}_n(\mathbf{x})]^2 d\mathbf{x}}{\int \hat{f}_m(\mathbf{x})^2 d\mathbf{x} + \int \hat{f}_n(\mathbf{x})^2 d\mathbf{x}}. \quad (25)$$

Since the mixture representation is non-parametric which makes the distance metric intractable to evaluate, we fit a Gaussian distribution to each component and obtain the approximated means $\hat{\mu}_m$ and variances $\hat{\Sigma}_m$ which are then substituted into (25). The distance metric is approximated by

$$dist(m,n) \approx \frac{|4\pi\hat{\Sigma}_m|^{-\frac{1}{2}} + |4\pi\hat{\Sigma}_n|^{-\frac{1}{2}} - 2\mathcal{N}(\hat{\mu}_m|\hat{\mu}_n, \Sigma_m + \Sigma_n)}{|4\pi\hat{\Sigma}_m|^{-\frac{1}{2}} + |4\pi\hat{\Sigma}_n|^{-\frac{1}{2}}}. \quad (26)$$

## VI. NUMERICAL CASE STUDY

In this section, we use a general numerical case to demonstrate the effectiveness and accuracy of the CMSMC-based tracking and prediction framework. Firstly, the superiority of CMSMC method is demonstrated by a comparison with EKF and UKF. Secondly, under the proposed framework we compare the recognition and tracking performance of the proposed HTSPM with widely used probabilistic classifiers and canonical state evolution models, respectively.

### A. Problem Formulation

In this example, we set a fixed tracking and prediction entity number and each entity is assigned one of the four high-level behaviors shown in Fig. 5. The training, validation and test trajectories for the $i$-th entity are generated by a nonlinear state space model with state constraints

$$x^i_{1,k} = x^i_{1,k-1} + 2x^i_{2,k-1}\Delta T + x^i_{3,k-1}\Delta T^2 + v^i_{1,k-1},$$
$$x^i_{2,k} = x^i_{2,k-1} + x^{i2}_{3,k-1}\Delta T + v^i_{2,k-1},$$
$$x^i_{3,k} = x^i_{3,k-1} + g^i(k-1) + v^i_{3,k-1},$$
$$z^i_{1,k} = x^i_{1,k} + w^i_{1,k}, \quad w^i_1 \sim \mathcal{N}(0, 0.5),$$
$$z^i_{2,k} = x^i_{2,k} + w^i_{2,k}, \quad w^i_2 \sim \mathcal{N}(0, 0.5),$$
$$x^i_{2,k} \geq 0, \quad -10 \leq x^i_{3,k} \leq 10,$$
$$x^i_{1,0} \sim U[0, 20], \quad x^i_{2,0} \sim U[10, 20], \quad x^i_{3,0} \sim \mathcal{N}(0, 0.1),$$
$$v^i_1 \sim \mathcal{N}(0, 0.5), \quad v^i_2 \sim U[-1, 1], \quad v^i_3 \sim U[-0.1, 0.1], \quad (27)$$

where $\Delta T$ is the period length between two time steps, $v_{j,k-1}$ ($j = 1, 2, 3$) is process noise, $w_k$ is measurement noise, $U[\cdot, \cdot]$ and $\mathcal{N}(\cdot, \cdot)$ denotes uniform and Gaussian distribution respectively and $g(k)$ is a manually defined function whose detailed forms for each sub-stages (1)-(6) are provided in Table I. This process model is nonlinear with non-Gaussian noise. The generated state trajectories of four behaviors can be spatially distinguished in the state space. We do not endow any physical meanings to states for generalization purpose.

### B. Experiment Details and Results

We demonstrate the validity and effectiveness of the proposed framework and models by the following experiments. The experiment details are provided and results are analyzed.

*1) CMSMC v.s. EKF and UKF:* To illustrate the advantages of proposed CMSMC method, we compared its tracking performance with EKF and UKF. Since linearization process is necessary for EKF, we adopted a differentiable state equation

$$x_{1,k} = x_{1,k-1} + 2x_{2,k-1}\Delta T + x_{3,k-1}\Delta T^2 + v_{1,k-1},$$
$$x_{2,k} = x_{2,k-1} + x^2_{3,k-1}\Delta T + v_{2,k-1},$$
$$x_{3,k} = x_{3,k-1} + v_{3,k-1}, \quad (28)$$

which is an approximation of the original state space model. We used 100 particles for each mixture component in CMSMC and there were four tracking targets corresponding to four different high-level behaviors. The performance comparisons are provided in Fig. 6. It is shown in Fig. 6(a) that the UKF and CMSMC have comparable accuracy on state mean values while EKF has a larger error especially during highly nonlinear stages, which indicates that using a first-order approximation at the current state (which is only employed in EKF) is not sufficient for estimating a general highly nonlinear system. Apart from mean values, covariance is another critical indicator when evaluating an approximated distribution. To be comparable with CMSMC, we sampled the same amount of particles using the estimated means and covariance matrices of EKF and UKF and computed the Mean Absolute Error (MAE) which is shown in Fig. 6(b). The results indicate that CMSMC can achieve the smallest MAE and the most stable tracking performance, which shows that CMSMC has greater advantages when handling nonlinear systems with non-Gaussian noise. The particles of CMSMC are visualized in Fig. 6(c) which achieves a smooth and coherent tracking performance.

*2) DHMM v.s. Other Probabilistic Classifiers:* Under this problem formulation, the DHMM has three layers corresponding to the three stage levels whose semantic labels are presented detailedly in Table II. Fig. 7 demonstrates the recognition results of each layer in DHMM for four cases in different behavior classes which possess a good interpretability. The comparisons of recognition results between DHMM and widely used probabilistic classifiers are shown in Fig. 8. Detailed analysis can be found in the captions.

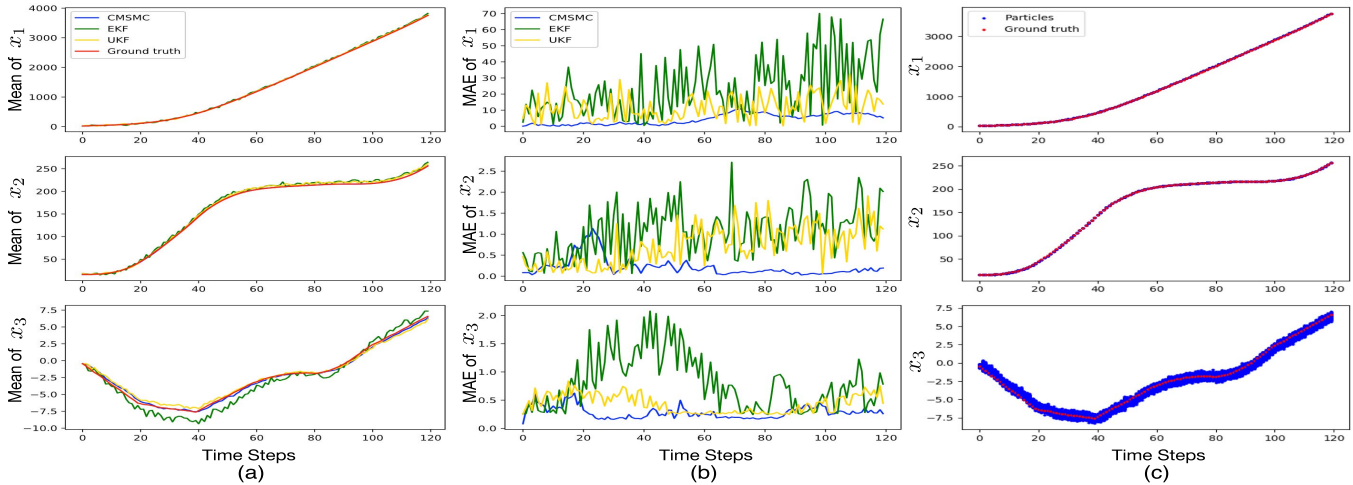

Fig. 6. The performance comparisons of CMSMC, EKF and UKF. (a) The mean values of state tracking results; (b) The mean absolute error (MAE) of state tracking results; (c) The visualized particles of CMSMC.

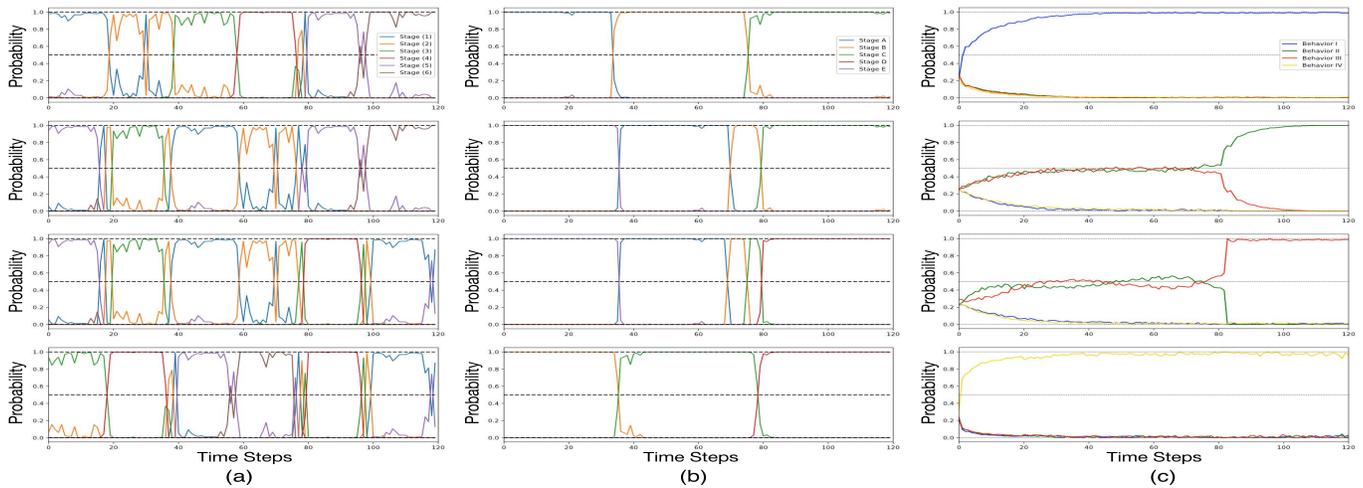

Fig. 7. The visualization for outputs of three-layer DHMM. Each row illustrates a case from a certain behavior. (a) The first layer output: the first-layer HMM can recognize six lowest-level stages and provide proper probability sequences despite some noise; (b) The second layer output: the second-layer HMM recognize five mid-level stages based on the output of first layer, which gives probability sequences with smaller noise; (c) The third layer output (behavior class probabilities): since there is an overlap of low-level stages between Behavior II and III, it is reasonable that the DHMM cannot distinguish them until the trajectories evolve differently, thus predicts a probability around 0.5 for each before divergence and recognize the right class quickly thereafter.

TABLE II
THE SEMANTIC LABELS OF DHMM (NUMERICAL EXAMPLE)

| Index | Stage | Index | Stage | Index | Stage |
|---|---|---|---|---|---|
| HMM-1-1 | (1) | HMM-1-6 | (6) | HMM-2-5 | E |
| HMM-1-2 | (2) | HMM-2-1 | A | HMM-3-1 | I |
| HMM-1-3 | (3) | HMM-2-2 | B | HMM-3-2 | II |
| HMM-1-4 | (4) | HMM-2-3 | C | HMM-3-3 | III |
| HMM-1-5 | (5) | HMM-2-4 | D | HMM-3-4 | IV |

*3) HTSPM v.s. Other State Evolution Models:* The HTSPM consists of the aforementioned DHMM and four independent state evolution models corresponding to four high-level behaviors I, II, III and IV. We compared the tracking performance of proposed HTSPM with other state evolution models based on the CMSMC framework in terms of the average of MAE values over the tracking horizon, which is presented in Table III. The *Global Gaussian Mixture Regression (GGMR), P-MLP* and *P-LSTM* are unified models which are trained without separating different behaviors, which means a single model is able to make predictions of all the behavior classes. The *SSM* refers to the aforementioned approximated state space model (28). For both conditional and unconditional models, the GMM has 20 mixture components and the neural network has three hidden layers each with 64 units followed by a ReLU activation function. The input noise is sampled from a three-dimensional normal distribution.

It can be seen that in general HTSPM can achieve lower MAE than the corresponding behavior-unconditional versions, which illustrates the significance of recognition module.

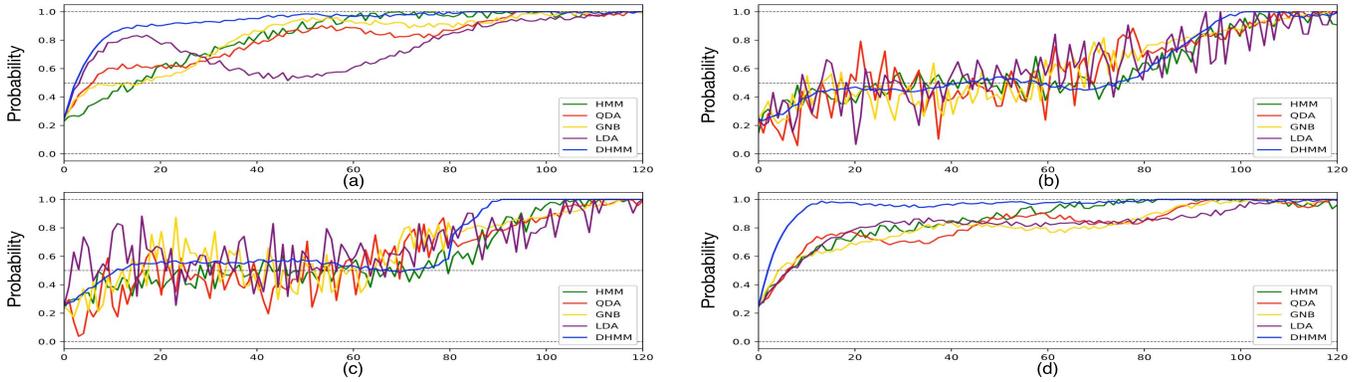

Fig. 8. The comparisons of recognition results of probabilistic classifiers for four behavior classes. (a) Behavior I case; (b) Behavior II case; (c) Behavior III case; (d) Behavior IV case. In Fig. 8(a) and 8(d), our DHMM is able to recognize the true behavior class earliest among the five classifiers as well as output a relatively stable probability. In Fig. 8(b) and 8(c), there exist much larger fluctuations and noise in the recognition results of the other classifiers than DHMM especially in the first 80 time steps during which Behavior II and III are identical, which indicates our DHMM is more robust to feature fluctuations and noise. The reason is that while the inference outputs of other classifiers are highly dependent on raw feature sequences, our model is able to extract different level meta feature sequences which are more stable and easier to classify, which reduces the influences of raw feature noise.

TABLE III
MAE VALUE COMPARISONS OF TRACKING PERFORMANCE FOR NUMERICAL CASE STUDY

| Model | $x_1$ | $x_2$ | $x_3$ |
|---|---|---|---|
| DHMM + CGMR | 1.43 | 0.51 | 0.71 |
| DHMM + CP-MLP | 1.15 | 0.38 | 0.57 |
| DHMM + CP-LSTM | **0.91** | **0.16** | **0.24** |
| GGMR | 2.98 | 0.74 | 1.12 |
| P-MLP | 1.67 | 0.61 | 0.88 |
| P-LSTM | 1.33 | 0.35 | 0.62 |
| SSM | 1.89 | 0.81 | 0.96 |

The reasons are two folds. On the one hand, the behavior-unconditional models need to learn a much more complex data distribution than behavior-conditional ones due to the variety of motion patterns, which demands a more sophisticated architecture with larger learning capacity. On the other hand, there tends to be mode collapse in behavior-unconditional models since the optimization algorithm usually gets stuck at local optimums and provide an averaged output of training cases which leads to undesired minimization of the loss functions. Among the compared models, using LSTM achieved the lowest tracking error which implies it is better at learning time dependencies of time-series. Moreover, note that overall learning-based models can achieve better tracking accuracy than the approximated state space model, which indicates high practicability and superiority of learning-based models in real-world applications where true system models are unavailable.

## VII. VEHICLE TRACKING AND PREDICTION

In this section, we apply the proposed framework and hierarchical time-series prediction model to solve real-time vehicle motion tracking and behavior prediction problems. We investigate a highway scenario as an illustrative example in which we only consider lane keeping and lane change behaviors due to the restriction of road geometry. The data source, experiment details, results and comparisons of different models are illustrated and discussed.

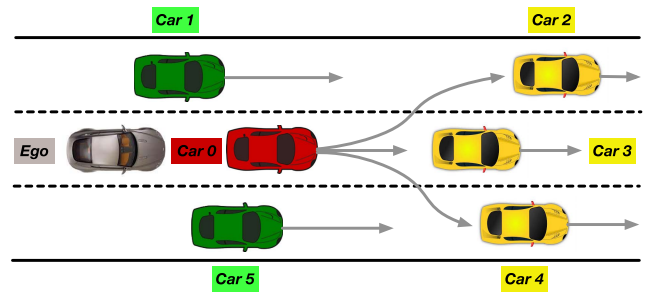

Fig. 9. A simplified representation of highway scenario. The gray car represents the ego autonomous vehicle with onboard sensors detecting its surrounding objects and the red car is object of study which may have interactions with the green ones and be affected by the motions of leading yellow ones.

### A. Problem Statement

We consider two observation perspectives: from the ego vehicle or from traffic surveillance systems. For the ego vehicle, the surrounding environment information is provided by onboard sensors which covers a certain range. It aims at tracking surrounding objects as well as forecasting their future behaviors. For surveillance systems, the traffic situations can be obtained by camera based monitors. Unlike the setup in numerical case, the number of tracking targets around the ego vehicle or within the monitor area may fluctuate as time goes by. Therefore, the adaptive mixture update mechanism is applied to fit the varying surrounding traffic situations (i.e. set $MUM = 1$ in Algorithm 1). For both tracking and prediction task, we make a reasonable simplification for situation representation with a group of six cars which is shown in Fig. 10, where we assume that only the red car can make a left lane change (LCL) or right lane change (LCR) while surrounding cars maintain the lane keeping behavior. This is a reasonable assumption since it is rare in realistic driving scenarios that two or more vehicles change lane simultaneously under this representation. Our goal is to track the six cars and make probabilistic predictions for their future behaviors.

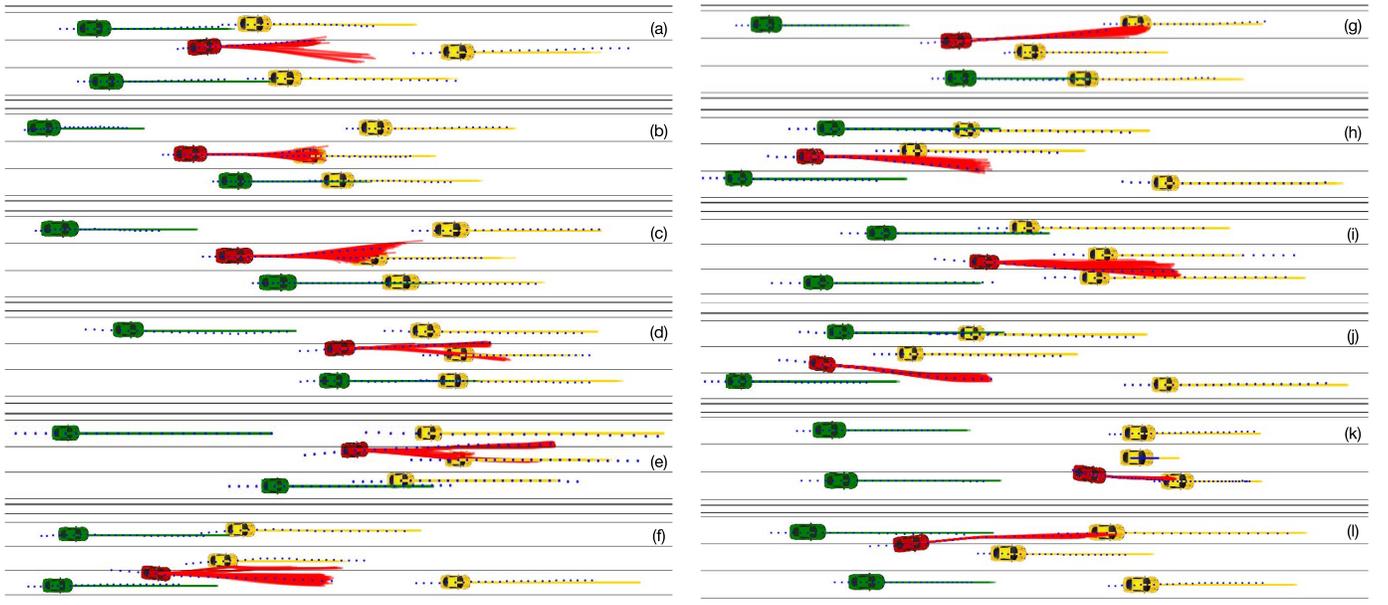

Fig. 10. Visualization of prediction results of selected typical cases of LCL and LCR behaviors using the *DHMM+CP-LSTM* model. The colored solid lines are sampled future trajectory hypotheses and the blue scatters are groundtruth trajectories.

### B. Data Source and Pre-Processing

The Next Generation Simulation (NGSIM) dataset is employed as the data source for extracting training, validation and test trajectories of vehicles, which can be found on [44]. The original dataset provides the estimated vehicle position, velocity, acceleration and other environment information extracted by image processing techniques from videos recording the traffic flow on a approximately 640 meters highway in California, USA. However, in some cases there is large detection noise and error especially on velocity and acceleration information which leads to unsmooth or unfeasible motions as indicated in [45]. Therefore, we applied an Extended Kalman Filter (EKF) to smooth and calibrate vehicle trajectories before experiments. We randomly selected 1,000 lane keeping cases and 200 lane change cases that satisfy the assumption in Section VII-A and split them into 70% as training data, 10% as validation data and 20% as test data.

### C. Vehicle Motion Models

In this work we investigate and compare two types of vehicle motion models: pure kinematic model and proposed hierarchical time-series model.

*1) Pure Kinematics Models:* In most of the vehicle tracking and motion prediction literature, kinematic models are naturally employed as the state transition model to propagate particle hypotheses. A comparison study of various motion models and their state transition equations are presented in [46]. The simplest models are *constant velocity model* (CVM) and *constant acceleration model* (CAM) which are linear models treating all 2D motions as translations in both longitudinal and lateral directions without considering rotations. More complicated models such as bicycle models also consider the yaw rate. However, the yaw rate can be assumed to be zero, which is reasonable in highway scenarios due to small yaw angle variations.

*2) Hierarchical Time-Series Prediction Model (HTSPM):* There are three high-level behaviors in the studied scenario. The recognition module is a two-layer DHMM whose semantic labels are introduced in Table IV. *Lane Keeping (LK)* behavior only consists of *Car Following* stage; *Lane Change Left (LCL)* and *Lane Change Right (LCR)* behaviors both consist of *Preparation*, *Deviation* and *Adjustment* stages successively. All of the six vehicles are taken into account to obtain the distribution of red car's future high-level behaviors.

The evolution module consists of three independent models corresponding to three behaviors respectively which are trained separately. Each behavior model forecasts motions of a portion of the six entities according to relevance. Specifically, the *LK* model considers *Car 0* and *Car 3* while the *LCL* model and *LCR* model consider *Car 0, Car 1, Car 2* and *Car 0, Car 4, Car 5*, respectively. The recognition module determines the proportion of sampled trajectories by each model in the evolution module.

TABLE IV
THE SEMANTIC LABELS OF DHMM (VEHICLE TRACKING AND PREDICTION)

| Index | Behavior | Stage |
|---|---|---|
| HMM-1-1 | Lane Keeping (LK) | Car following |
| HMM-1-2 | Lane Change Left (LCL) | Preparation |
| HMM-1-3 | Lane Change Left (LCL) | Deviation |
| HMM-1-4 | Lane Change Left (LCL) | Adjustment |
| HMM-1-5 | Lane Change Right (LCR) | Preparation |
| HMM-1-6 | Lane Change Right (LCR) | Deviation |
| HMM-1-7 | Lane Change Left (LCL) | Adjustment |
| HMM-2-1 | Lane Keeping (LK) | — |
| HMM-2-2 | Lane Change Left (LCL) | — |
| HMM-2-3 | Lane Change Right (LCR) | — |

*3) Behavior-Unconditional Learning Based Models:* Behavior-unconditional models treat all the entities as a whole system and predict the joint distribution of their motions. The objective is thus to capture the multi-modality of data distribution raised by multiple behavior patterns, which is much harder to achieve due to the demand for large representation capacity. To allow for a fair comparison, these models should have more complicated architectures than the evolution module of HTSPM.

### D. Experiments Details and Results

For the recognition module of HTSPM, we decided the number of hidden states of each HMM according to BIC score. Since the Baum-Welch algorithm can reach different solutions with diverse initialization, we trained each HMM multiple times and selected the model with highest final BIC score. For the evolution module, the CGMR has 10 mixture components; the CP-MLP consists of four hidden fully-connected layers with 64 units followed by a leaky ReLU activation function; the CP-LSTM has the same architecture as CP-MLP except that the first fully-connected layer is replaced with a LSTM layer. For the behavior-unconditional models we increased the model capacity. The GGMR has 30 mixture components; the P-MLP and P-LSTM also consist of four hidden layers but with 192 units. We trained all the models multiple times and selected the ones with smallest prediction error on the validation set. Moreover, we chose CAM from vehicle kinematic models as a baseline.

We adopted a unified input feature representation for models of the same type. Specifically, for behavior-conditional models (i.e. *CGMR, CP-MLP* and *CP-LSTM*) the feature contains a sequence of historical relative positions of only model-related surrounding vehicles with respect to the middle red vehicle $\{x_i(k-T:k), y_i(k-T:k), i = 0, \ldots, 5\}$ as well as their absolute advancing velocities $\{v_i(k-T:k-1), i = 0, \ldots, 5\}$, where $T$ is history horizon; while for behavior-unconditional models (i.e. *GGMR, GP-MLP* and *GP-LSTM*), the feature covers the same information of all the six vehicles.

*1) Quantitative Analysis:* For multi-target tracking, we sampled 100 initial particles for each tracked vehicle from a Gaussian distribution with the mean at initial observations. The particle state contains vehicle positions $x$, $y$ and velocities $\dot{x}$, $\dot{y}$ except that accelerations $\ddot{x}$, $\ddot{y}$ are additionally considered when using CAM. The comparisons of tracking performance in terms of Average Distance Error (ADE) are illustrated in Table V where the first column corresponds to middle car while the second one corresponds to the average of five surrounding cars. The bold numbers indicate best performance. It is shown that learning-based vehicle motion models can achieve much lower tracking errors than pure kinematic model. Moreover, with the behavior recognition module on top of evolution module, the HTSPM is more capable of capturing the true vehicle state evolution distribution.

TABLE V
ADE VALUE COMPARISONS OF TRACKING PERFORMANCE OF VEHICLE POSITIONS AND VELOCITIES

| Model | Position (m) | Velocity (m/s) |
|---|---|---|
| DHMM + CGMR | 0.046 / 0.028 | 0.636 / 0.541 |
| DHMM + CP-MLP | 0.041 / **0.021** | **0.547** / 0.488 |
| DHMM + CP-LSTM | **0.038** / 0.024 | 0.582 / **0.473** |
| GGMR | 0.051 / 0.033 | 0.941 / 0.557 |
| GP-MLP | 0.043 / 0.029 | 0.812 / 0.617 |
| GP-LSTM | 0.044 / 0.031 | 0.847 / 0.656 |
| CAM | 0.082 / 0.067 | 1.936 / 1.614 |

For multi-agent prediction, we sampled 100 particle hypotheses for each entity to make predictions in all the experiments. Table VI provides the ADE value comparisons of vehicle position prediction using both HTSPM and baseline models. It is shown that employing proposed HTSPM can achieve the lowest prediction error. The *DHMM+CP-LSTM* has superiority over the others in most time steps, which implies that recurrent neural network is more capable of learning long-term dependencies. Although the *CAM* can achieve acceptable performance in the first second, the error increases greatly as prediction horizon expands, which indicates pure kinematic models are only suitable for short-term predictions.

*2) Qualitative Analysis:* The prediction results of several test cases are visualized in Fig. 10 to demonstrate the model performance. In the first column of figures, it can be seen that our model is able to make multi-modal trajectory predictions considering uncertainty both on the behavior and motion level. In Fig. 10(a), the red car is predicted to make a left lane change according to its heading tendency. However, due to the small gap between the green and yellow cars on the target lane and their relative velocities, it is also possible and reasonable for the red car to change its mind to continue the lane keeping behavior, which is also captured by the proposed model. Fig. 10(b) and 10(c) show two chronological time steps in the same test case where at first the red car may choose all three possible behaviors while after a moment the probability of LCL increases due to the large gap on its left, small gap on its right and low relative velocity of its leading car, which demonstrates the online evolution of prediction results. The second column of figures mainly show the later stage of lane change where only one of the three behaviors dominates the future trajectories, where the variance of samples is still maintained to present different driving patterns.

*3) Ablative Analysis:* We also conducted an ablative analysis to demonstrate the relative importance of recognition and evolution module through comparing the prediction errors under four model settings:

*a) GT + behavior-conditional model:* We directly use the state evolution model corresponding to the groundtruth behavior. This can be treated as an upper limit of prediction performance of evolution module.

*b) DHMM + behavior-conditional model:* This is just to use the complete proposed HTSPM with state constraints.

TABLE VI
ADE VALUE COMPARISONS OF VEHICLE POSITION PREDICTION

| Cases | Prediction Horizon (s) | DHMM + CGMR (m) | DHMM + CP-MLP (m) | DHMM + CP-LSTM (m) | GGMR (m) | GP-MLP (m) | GP-LSTM (m) | CAM (m) |
|---|---|---|---|---|---|---|---|---|
| Lane Keeping (LK) | 1.0 | **0.12** / **0.11** | 0.21 / 0.20 | 0.18 / 0.21 | 0.19 / 0.25 | 0.33 / 0.30 | 0.32 / 0.31 | 0.41 / 0.39 |
|  | 2.0 | 0.93 / 0.99 | **0.68** / 0.65 | 0.72 / **0.62** | 1.01 / 0.97 | 0.80 / 0.74 | 0.76 / 0.74 | 1.91 / 1.33 |
|  | 3.0 | 2.21 / 1.67 | 1.14 / **1.36** | **1.13** / 1.38 | 2.33 / 1.69 | 1.26 / 1.48 | 1.29 / 1.42 | 3.36 / 2.90 |
|  | 4.0 | 3.37 / 2.29 | 1.69 / **2.20** | **1.65** / 2.31 | 3.68 / 2.34 | 1.81 / 2.32 | 1.78 / 2.29 | 4.65 / 3.63 |
|  | 5.0 | 4.05 / 3.83 | 2.91 / 3.33 | **2.90** / **3.04** | 4.41 / 3.88 | 3.03 / 3.45 | 3.14 / 3.28 | 5.88 / 5.02 |
| Lane Change Left (LCL) | 1.0 | 0.36 / **0.13** | 0.31 / 0.22 | **0.29** / 0.24 | 0.42 / 0.28 | 0.41 / 0.34 | 0.38 / 0.35 | 0.51 / 0.42 |
|  | 2.0 | 0.99 / 1.06 | 0.96 / 0.87 | **0.91** / **0.83** | 1.15 / 1.04 | 1.11 / 0.99 | 1.14 / 0.98 | 2.39 / 1.51 |
|  | 3.0 | 1.98 / 1.79 | 1.79 / 1.82 | **1.69** / **1.57** | 2.11 / 1.93 | 1.91 / 1.94 | 1.83 / 1.90 | 3.73 / 5.64 |
|  | 4.0 | 2.89 / 2.86 | **2.61** / **2.41** | 2.91 / 2.69 | 3.26 / 2.94 | 2.73 / 2.98 | 2.71 / 2.98 | 6.65 / 3.77 |
|  | 5.0 | 4.63 / 4.23 | 3.82 / 3.78 | **3.74** / **3.58** | 5.37 / 4.11 | 4.04 / 3.90 | 3.95 / 3.73 | 7.68 / 5.36 |
| Lane Change Right (LCR) | 1.0 | 0.29 / 0.23 | 0.27 / **0.18** | **0.24** / 0.25 | 0.49 / 0.45 | 0.38 / 0.35 | 0.35 / 0.37 | 0.59 / 0.46 |
|  | 2.0 | **0.83** / 1.12 | 1.02 / 0.86 | 0.97 / **0.81** | 0.91 / 1.08 | 1.14 / 0.97 | 1.09 / 0.92 | 2.23 / 1.49 |
|  | 3.0 | 1.94 / 1.66 | 1.73 / 1.84 | **1.68** / **1.64** | 2.15 / 1.68 | 1.83 / 1.95 | 1.79 / 1.98 | 3.58 / 2.71 |
|  | 4.0 | 3.01 / 2.52 | 2.58 / **2.41** | **2.44** / 2.56 | 3.28 / 2.82 | 2.74 / 2.67 | 2.78 / 2.70 | 6.23 / 3.68 |
|  | 5.0 | 4.58 / 3.92 | 3.76 / **3.61** | **3.71** / 3.82 | 5.05 / 4.31 | 3.91 / 3.82 | 3.84 / 3.79 | 7.21 / 5.54 |

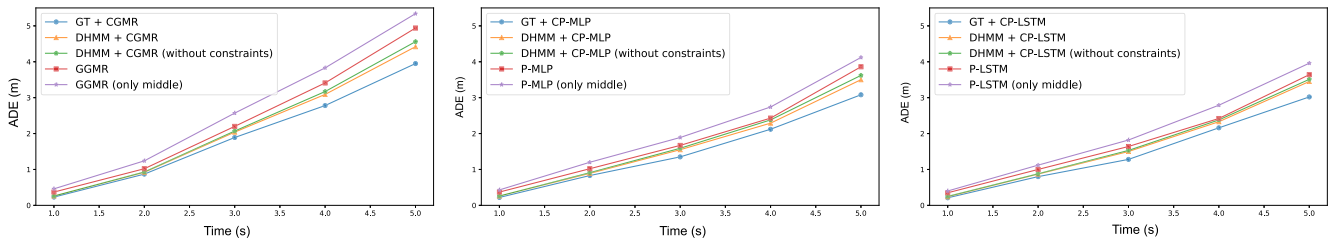

Fig. 11. The ablative analysis of prediction accuracy of middle vehicle position in terms of Average Distance Error (ADE) using different learning-based state evolution models with different recognition results.

*c) DHMM + behavior-conditional model (no constraints):* This is to use the proposed HTSPM but without considering constraints on the vehicle state.

*d) Behavior-unconditional model:* This is the learning-based baseline model without classification on behaviors.

*e) Behavior-unconditional model (only middle car):* This model does not consider the surrounding vehicles and make predictions for the middle car with only its historical trajectories, which is used to illustrate the significance of considering interactions among entities.

Fig. 11 shows the ADE values of prediction for the above model settings. We find that models considering adjacent vehicles outperforms those only focusing on the middle vehicle, implying that the motions of surrounding cars have significant influence on the target vehicle. The *DHMM+Behavior-conditional Model* leads to further improvement, suggesting the effectiveness of behavior recognition prior to motion forecasting. Both factors become more remarkable as prediction horizon extends. It is also shown that incorporating kinematic constraints on vehicle state can achieve better prediction accuracy than otherwise despite that the improvement is not significant, which indicates that there are not many violations of constraints in the output action of proposed models. Moreover, from the performance of groundtruth behavior model we observe some space for improvement if the recognition module becomes more powerful.

## VIII. CONCLUSION

In this paper, a generic multi-target tracking and multi-agent probabilistic behavior prediction framework based on constrained mixture sequential Monte Carlo (CMSMC) method was proposed, which can track multiple entities simultaneously without explicit data association with a unified representation and predict the joint distribution of their future motions or states. A generic learning-based hierarchical time-series prediction model (HTSPM) was also put forward to serve as an implicit proposal distribution in the prior update of Bayesian state estimation. The proposed framework and models were applied to a numerical case study and a real-world on-road vehicle tracking and behavior prediction task in highway scenarios. The results show that the proposed CMSMC method can achieve better tracking accuracy than variants of KF in terms of both mean and variance of posterior distribution. The DHMM in recognition module of HTSPM can better capture the behavior distribution than other probabilistic classifiers in terms of response time and robustness. Multiple state evolution models including learning-based ones and pure kinematics-based ones were compared under the framework settings. An ablative analysis was also conducted to demonstrate the significance of constraint incorporation and recognition module. Future research directions include enhancing the capability of both recognition module and evolution module and applying the proposed framework to more complicated scenarios with more interactions and mutual reactions such as roundabout and unsignalized intersections.

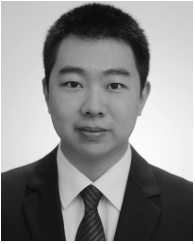

**Jiachen Li** received the B.E. degree from the Control Science and Engineering Department, Harbin Institute of Technology, China, in 2016. He is currently pursuing the Ph.D. degree with the Department of Mechanical Engineering, University of California at Berkeley, Berkeley, USA. He was a Research Assistant with the Research Institute of Intelligent Control and Systems, Harbin Institute of Technology, from 2014 to 2016. His research interests include machine learning, optimization, computer vision approaches and their applications to behavior prediction, and decision making and motion planning for multi-agent intelligent systems such as autonomous vehicles and robotics.

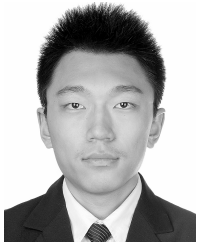

**Wei Zhan** received the B.E. and M.S. degrees from the Control Science and Engineering Department, Harbin Institute of Technology, Harbin, China, in 2012 and 2014, respectively. He is currently pursuing the Ph.D. degree with the Mechanical Engineering Department, University of California at Berkeley, Berkeley, CA, USA. He was a Research Assistant with the Department of Mechanical Engineering, The University of Hong Kong, in 2012. His research interests include decision-making, motion planning, motion prediction, and behavior analysis for autonomous driving.

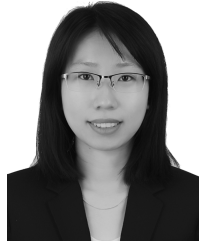

**Yeping Hu** received the B.S. degree in mechanical engineering from the University of Illinois at Urbana–Champaign, IL, USA, in 2016. She is currently pursuing the Ph.D. degree with the Department of Mechanical Engineering, University of California at Berkeley, Berkeley, CA, USA. She is a member of the Mechanical Systems Control Laboratory. Her research interests include behavior prediction, decision making, and motion planning for autonomous driving.

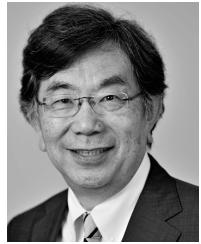

**Masayoshi Tomizuka** (M'86–SM'95–F'97–LF'16) received the B.S. and M.S. degrees in mechanical engineering from Keio University, Tokyo, Japan, and the Ph.D. degree in mechanical engineering from the Massachusetts Institute of Technology, in 1974.

In 1974, he joined the Faculty of the Department of Mechanical Engineering at the University of California at Berkeley (UC Berkeley), where he currently holds the Cheryl and John Neerhout, Jr., Distinguished Professorship Chair. At UC Berkeley, he teaches courses in dynamic systems and controls. His current research interests include optimal and adaptive control, digital control, motion control, and control problems related to robotics, precision motion control and vehicles. He served as a Program Director of the Dynamic Systems and Control Program of the Civil and Mechanical Systems Division of NSF from 2002 to 2004.

Dr. Tomizuka was a recipient of the Charles Russ Richards Memorial Award (ASME, 1997), the Rufus Oldenburger Medal (ASME, 2002), the John R. Ragazzini Award (AACC, 2006), Lifetime Achievement Award, Technical Committee on Mechatronic Systems, IFAC (2013), and the Richard E. Bellman Control Heritage Award (AACC, 2018). He was the General Chairman of the 1995 American Control Conference, and served as the President of the American Automatic Control Council (AACC) from 1998 to 1999. He served as a Technical Editor of the *ASME Journal of Dynamic Systems, Measurement and Control*, J-DSMC from 1988 to 1993, an Editor-in-Chief of the IEEE/ASME TRANSACTIONS ON MECHATRONICS from 1997 to 1999, and an Associate Editor of the *Journal of the International Federation of Automatic Control* (IFAC) and *Automatica*. He is a Life Fellow of the ASME.